\documentclass[final]{cvpr}

\usepackage{times}
\usepackage{epsfig}
\usepackage{graphicx}
\usepackage{amsmath}

\usepackage{amssymb}

\usepackage{subcaption}
\usepackage{multirow}

 \usepackage{paralist}
\usepackage{wrapfig}
\usepackage{booktabs}
\usepackage{makecell}

\usepackage{arydshln}
\usepackage{dsfont}
\usepackage{color}
\usepackage[font={footnotesize}]{subcaption}
\makeatletter
\@namedef{ver@everyshi.sty}{}
\makeatother
\usepackage{tikz}
\usepackage{pgfplots}
\pgfplotsset{compat=1.13}

\usepackage[font={footnotesize}]{caption}
\makeatletter
\g@addto@macro\normalsize{%
	\setlength\abovedisplayskip{5pt}
	\setlength\belowdisplayskip{2pt}
	\setlength\abovedisplayshortskip{2pt}
	\setlength\belowdisplayshortskip{2pt}
}
\makeatother
\definecolor{turquoise}{cmyk}{0.65,0,0.1,0.1}
\definecolor{purple}{rgb}{0.65,0,0.65}
\definecolor{dark_green}{rgb}{0, 0.5, 0}
\definecolor{orange}{rgb}{0.8, 0.6, 0.2}
\definecolor{red}{rgb}{0.8, 0.2, 0.2}
\definecolor{brown}{rgb}{0.5, 0.16, 0.16}

\newcommand{\tbd}[1]{{\color{red}TBD}}

\newcommand{\tr}{{\mathsf{T}}}

\renewcommand{\vec}[1]{\ensuremath{\mathbf{#1}}}
\newcommand{\q}{\vec{q}}
\newcommand{\p}{\vec{p}}

\newcommand{\n}{\vec{n}}
\newcommand{\R}{\mathds{R}}
\renewcommand{\P}{\mathcal{P}}
\newcommand{\Q}{\mathcal{Q}}
\renewcommand{\S}{\mathcal{S}}
\newcommand{\N}{\mathcal{N}}

\usepackage[pagebackref=true,breaklinks=true,colorlinks,bookmarks=false]{hyperref}

\def\cvprPaperID{5444} 
\def\confYear{CVPR 2021}

\begin{document}

\title{Iso-Points: Optimizing Neural Implicit Surfaces with Hybrid Representations}

\author{
	Wang Yifan\textsuperscript{1}\hspace{1.0em} 
	Shihao Wu\textsuperscript{1}\hspace{1.0em}  
	Cengiz {\"O}ztireli\textsuperscript{2}\hspace{1.0em} 
	Olga Sorkine-Hornung\textsuperscript{1}\hspace{1.0em}
	\\\\
	\textsuperscript{1}ETH Zurich \hspace{1.0em} \textsuperscript{2}University of Cambridge
	\vspace{-8pt}
}

\maketitle

\begin{abstract}
Neural implicit functions have emerged as a powerful representation for surfaces in 3D. Such a function can encode a high quality surface with intricate details into the parameters of a deep neural network. However, optimizing for the parameters for accurate and robust reconstructions remains a challenge,  especially when the input data is noisy or incomplete. In this work, we develop a hybrid neural surface representation that allows us to impose geometry-aware sampling and regularization, which significantly improves the fidelity of reconstructions. We propose to use \emph{iso-points} as an explicit representation for a neural implicit function. These points are computed and updated on-the-fly during training to capture important geometric features and impose geometric constraints on the optimization. We demonstrate that our method can be adopted to improve state-of-the-art techniques for reconstructing neural implicit surfaces from multi-view images or point clouds. Quantitative and qualitative evaluations show that, compared with existing sampling and optimization methods, our approach allows faster convergence, better generalization, and accurate recovery of details and topology.

\end{abstract}

\vspace{-0.5cm}
\section{Introduction}\label{sec:intro}

\begin{figure}\centering
\includegraphics[width=0.9\linewidth,trim={0 8mm 0 0}]{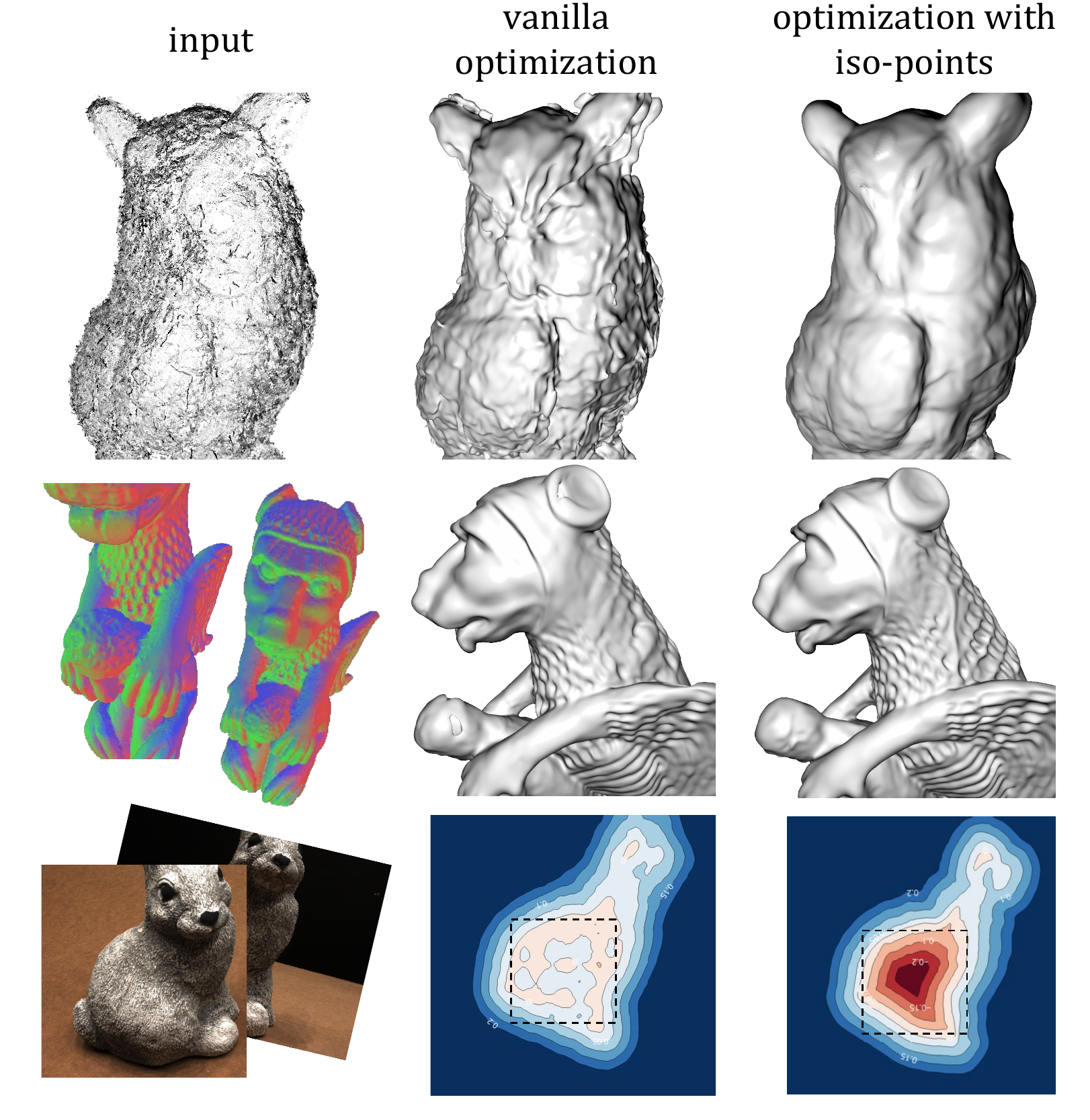}
\caption[caption for teaser]{We propose a hybrid neural surface representation with implicit functions and \emph{iso-points}, which leads to accurate and robust surface reconstruction from imperfect data. 
The iso-points allows us to augment existing optimization pipelines in a variety of ways: 
geometry-aware regularizers are incorporated to reconstruct a surface from a noisy point cloud (first row); geometric details are preserved in multi-view reconstruction via feature-aware sampling (second row); iso-points can serve as a 3D prior to improve the topological accuracy of the reconstructed surface (third row). 
The input data are respectively: reconstructed point cloud \cite{furukawa2009accurate} of model 122 of DTU-MVS dataset~\cite{jensen2014large}, multi-view rendered images of \textsc{dog-winged} model from Sketchfab dataset \cite{yifan2019patch} and multi-view images of model 55 of DTU-MVS dataset}\label{fig:teaser}
\end{figure}

Reconstructing surfaces from real-world observations is a long-standing task in computer vision.
Recently, representing 3D geometry as a neural implicit function has received much attention~\cite{park2019deepsdf,mescheder2019occupancy,chen2019learning}. Compared with other 3D shape representations, such as point clouds~\cite{lin2017learning,insafutdinov2018unsupervised,yifan2019differentiable}, polygons~\cite{kato2018neural,liu2018paparazzi,nimier2019mitsuba,liu2019soft}, and voxels~\cite{choy20163d,tulsiani2017multi,lombardi2019neural,jiang2020sdfdiff}, it provides a versatile representation with infinite resolution and unrestricted topology.

Fitting a neural implicit surface to input observations is an optimization problem. Some common application examples include surface reconstruction from point clouds and multi-view reconstruction from images. For most cases, the observations are noisy and incomplete. This leads to fundamental geometric and topological problems in the final reconstructed surface, as the network overfits to the imperfect data. We observe that this problem remains, and can become more prominent with the recent powerful architectures, \eg sine activations \cite{sitzmann2019scene} and Fourier features \cite{pharr2016physically}.

We show examples of problems in fitting neural implicit functions in Figure~\ref{fig:teaser}. When fitting a neural surface to a noisy point cloud, ``droplets" and bumps emerge where there are outlier points and holes (first row); when fitting a surface to image observations, fine-grained geometric features are not captured due to under-sampling (second row); topological noise is introduced when inadequate views are available for reconstruction (third row). 

In this work, we propose to alleviate these problems by introducing a \emph{hybrid neural surface representation} using \emph{iso-points}. The technique converts from an implicit neural surface to an explicit one via sampling iso-points, and goes back to the implicit representation via optimization. The two-way conversion is performed on-the-fly during training to introduce geometry-aware regularization and optimization. This approach unlocks a large set of fundamental tools from geometry processing to be incorporated for accurate and robust fitting of neural surfaces.

A key challenge is to extract the iso-points on-the-fly efficiently and flexibly during the training of a neural surface. Extending several techniques from point-based geometry processing, we propose a multi-stage strategy, consisting of projection, resampling, and upsampling. We first obtain a sparse point cloud on the implicit surface via projection, then resample the iso-points to fix severely under-sampled regions, and finally upsample to obtain a dense point cloud that covers the surface. As all operations are GPU friendly and the resampling and upsampling steps require only local point distributions, the entire procedure is fast and practical for training.

We illustrate the utility of the new representation with a variety of applications, such as multi-view reconstruction and surface reconstruction from noisy point clouds.
Quantitative and qualitative evaluations show that our approach allows for fast convergence, robust optimization, and accurate reconstruction of details and topology.

\section{Related work}
\label{related work}
We begin by discussing existing representations of implicit surfaces, move on to the associated optimization and differentiable rendering techniques given imperfect input observations, and finally review methods proposed to sample points on implicit surfaces.

\vspace{1mm}
\textbf{Implicit surface representations.}
Implicit functions are a flexible representation for surfaces in 3D. Traditionally, implicit surfaces are represented globally or locally with radial basis functions (RBF)~\cite{carr2001reconstruction}, moving least squares (MLS)~\cite{levin1998approximation}, volumetric representation over uniform grids~\cite{curless1996volumetric}, or adaptive octrees~\cite{kazhdan2006poisson}.
Recent works investigate neural implicit surface representations, i.e., using deep neural networks to encode implicit function~\cite{park2019deepsdf,sitzmann2019scene}, which achieves promising results in reconstructing surfaces from 3D point clouds~\cite{atzmon2020sal,sitzmann2020implicit,erler2020points2surf} or images~\cite{lin2020sdf,yariv2020universal,niemeyer2020differentiable}.

Compared with simple polynomial or Gaussian kernels, implicit functions defined by nested activation functions, e.g., MLPs~\cite{chen2019learning} or \textsc{SIREN}~\cite{sitzmann2019scene}, have more capability in representing complex structures. However, the fiting of such neural implicit function requires clean supervision points~\cite{xu2020ladybird} and careful optimization to prevent either overfitting to noise or underfitting to details and structure.

\vspace{1mm}
\textbf{Optimizing neural implicit surfaces with partial observations.}
Given raw 3D data, Atzmon et al.~\cite{atzmon2020sal,atzmon2020sald} use sign agnostic regression to learn neural implicit surfaces without using a ground truth implicit function for supervision. Gropp et al.~\cite{gropp2020implicit} use the Eikonal term for implicit geometric regularization and provide a theoretical analysis of the plane reproduction property possessed by the neural zero level set surfaces. Erler et al.~\cite{erler2020points2surf} propose a patch-based framework that learns both the local geometry and the global inside/outside information, which outperforms existing data-driven methods. None of these methods exploit an explicit sampling of the implicit function to improve the optimization. Poursaeed et al.~\cite{poursaeed2020coupling}  use two different encoder-decoder networks to simultaneously predict both an explicit atlas~\cite{groueix2018papier} and an implicit function. In contrast, we propose a hybrid representation using a single network.

When the input observations are in the form of 2D images, differentiable rendering allows us to use 2D pixels to supervise the learning of 3D implicit surfaces through automatic differentiation and approximate gradients~\cite{kato2020differentiable,tewari2020state}.
The main challenge is to render the implicit surface and compute reliable gradients at every optimization step efficiently. 
Liu et al.~\cite{liu2020dist} accelerate the ray tracing process via a coarse-to-fine sphere tracing algorithm \cite{hart1996sphere}, and use an approximate gradient in back propagation. 
In~\cite{liu2019learning}, a ray-based field probing and an importance sampling technique are proposed for efficient sampling of the object space. 
Although these methods greatly improve rendering efficiency, the sampling of ray-based algorithms, i.e., the intersection between the ray and the iso-surface, are intrinsically irregular and inefficient.
Most of the above differentiable renderers use ray casting to generate the supervision points.  We propose another type of supervision points by sampling the implicit surface in-place.

\vspace{1mm}
\textbf{Sampling implicit surfaces.}
In 1992, Figueiredo et al.~\cite{de1992physically} proposed a powerful way to sample implicit surfaces using dynamic particle systems that include attraction and repulsion forces.
Witkin and Heckbert~\cite{witkin1994using} further developed this concept by formulating an adaptive repulsion force.
While the physical relaxation process is expensive, better initialization techniques have been proposed, such as using seed flooding on the partitioned space~\cite{levet2006fast} or the octree cells~\cite{proencca2007sampling}.
Huang et al.~\cite{huang2013edge} resample point set surfaces to preserve sharp features by pushing points away from sharp edges before upsampling.
When sampling a neural implicit surface, existing works such as Atzmon et al.~\cite{atzmon2019controlling} project randomly generated 3D points onto the iso-surface along with the gradient of the neural level set. However, such sampled points are unevenly distributed, and may leave parts of the surface under-sampled or over-sampled.

\vspace{-1mm}
\section{Method}\label{sec:method}
Given a neural implicit function $ f_t(\p;\theta_t)$ representing the surface $ \S_t $, where $ \theta_t $ are the network parameters at the \hbox{$t$-th} training iteration, our goal is to efficiently generate and utilize a dense and uniformly distributed point set on the zero level set, called \emph{iso-points}, which faithfully represents the current implicit surface $ \S_t $.
Intuitively, we can deploy the iso-points back into the ongoing optimization to serve various purposes, \eg improving the sampling of training data and providing regularization for the optimization, leading to a substantial improvement in the convergence rate and the final optimization quality.

In this section, we first focus on how to extract the iso-points via projection and uniform resampling. We then explain how to utilize the iso-points for better optimization in practical scenarios.

\subsection{Iso-surface sampling}
\begin{figure}[t]
	\includegraphics[width=\linewidth, trim={0 0.5cm 0 0}, clip]{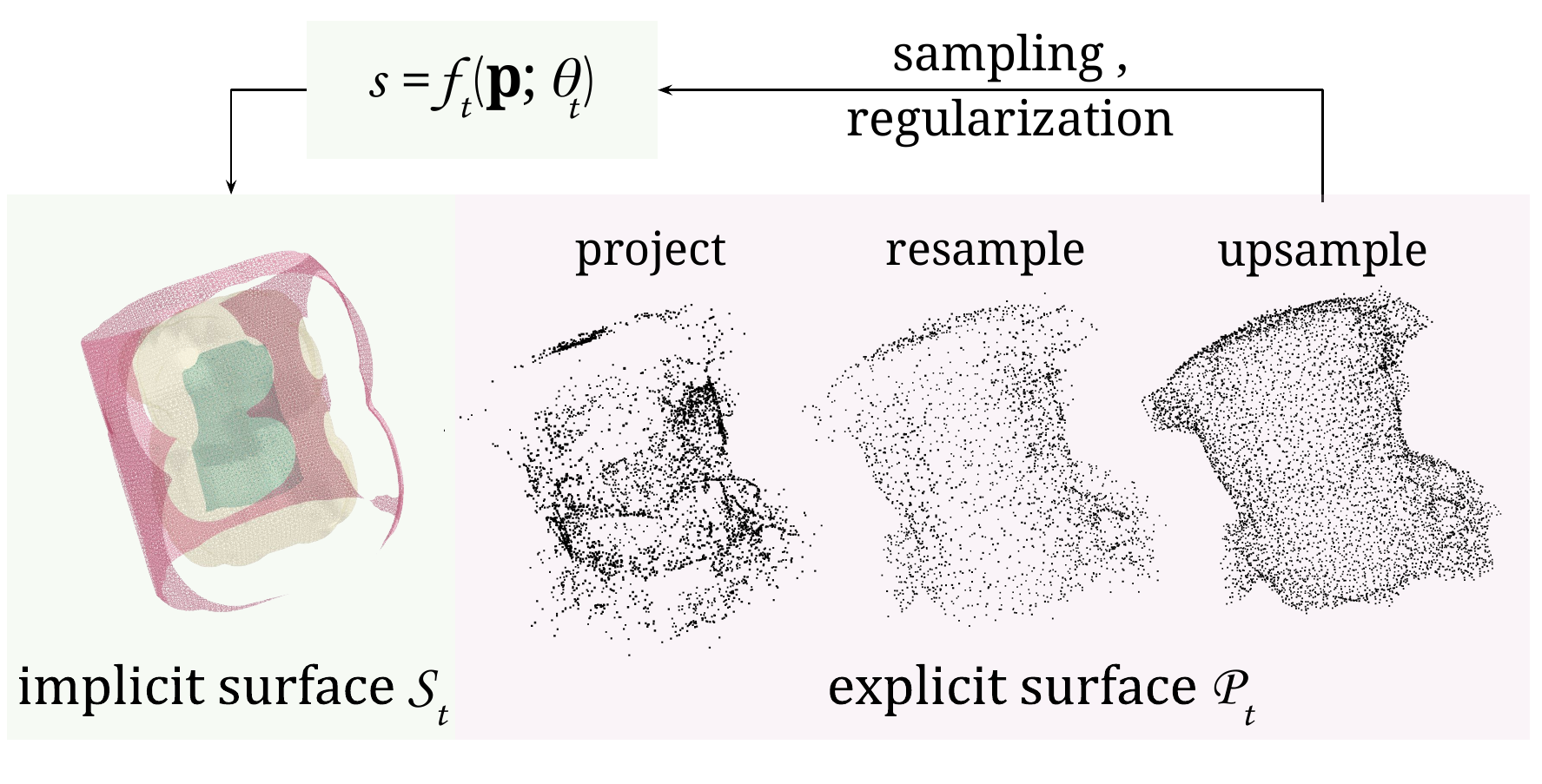}
	\caption{Overview of our hybrid representation. We efficiently extract a dense, uniformly distributed set of \emph{iso-points} as an explicit representation for a neural implicit surface. Since the extraction is fast, iso-points can be integrated back into the optimization as a 3D geometric prior, enhancing the optimization.}\label{fig:overview}
\end{figure}

As shown in Figure~\ref{fig:overview}, our iso-surface sampling consists of three stages. 
First, we project a set of initial points $ \Q_t $ onto the zero level set to get a set of \emph{base iso-points} $ \tilde{\Q}_t $.
We then resample $ \tilde{\Q}_t $ to avoid clusters of points and fill large holes.
Finally, we upsample the points to obtain dense and uniformly distributed iso-points $ \P_t $.

\vspace{2mm}
\textbf{Projection.} Projecting a point onto the iso-surface can be seen as using Newton's method~\cite{ben1966newton} to approximate the root of the function starting from a given point.
Atzmon \etal~\cite{atzmon2019controlling} derive the projection formula for functions modeled by generic networks.
For completeness, we recap the steps here, focusing on $ f:\R^3\to\R $.

Given an implicit function $ f(\p) $ representing a signed distance field and an initial point \hbox{$ \q_0 \in \R^3$}, we can find an iso-point $ \p $ on the zero level set of $f$ using Newton iterations:
\hbox{$ \q_{k+1} = \q_{k} - J_f(\q_{k})^{+}f(\q_{k})$,}  where $ J_f^{+}$ is the Moore-Penrose pseudoinverse of the Jacobian. 
In our case, $ J_f$ is a row $3$-vector, so that $  J_f^{+}(\q_{k}) =\frac{J_f^\tr(\q_{k})}{\|J_f(\q_{k})\|^2} $,
where the Jacobian $ J_f(\q_{k}) $ can be conveniently evaluated in the network via backpropagation.

However, for some contemporary network designs, such as sine activation functions~\cite{sitzmann2020implicit} and positional encoding~\cite{mildenhall2020nerf}, the signed distance field can be very noisy and the gradient highly non-smooth.
Directly applying Newton's method then causes overshooting and oscillation. 
While one could attempt more sophisticated line search algorithms, we instead address this issue with a simple clipping operation to bound the length of the update, \ie
\begin{equation}\label{eq:projection}\textstyle
	\q_{k+1} = \q_{k} - \tau\left(\frac{J_f^\tr(\q_{k})}{\|J_f(\q_{k})\|^2}f(\q_{k})\right),
\end{equation}
where $ \tau(\vec{v}) = \frac{\vec{v}}{\|\vec{v}\|}\min(\|\vec{v}\|, \tau_0) $. 
We set $ \tau_0=\frac{D}{2|\Q_t|} $ with $ D $ denoting the diagonal length of the shape's bounding box.

In practice, we initialize $ \Q_t $ with randomly sampled points at the beginning of the training and then with iso-points $ \P_{t-1} $ from the previous training iteration.
Similar to \cite{atzmon2019controlling}, at each training iteration, we perform a maximum of 10 Newton iterations and terminate as soon as all points have converged, \ie $ |f(\q_{k})| < \epsilon, \forall \q \in \Q_t$. The termination threshold $ \epsilon $ is set to $ 10^{-4}  $ and gradually reduced to $10^{-5} $ during training.

\vspace{1mm}
\textbf{Uniform resampling.}
The projected base iso-points $ \tilde{\Q}_t $ can be sparse and hole-ridden due to the irregularity present in the neural distance field, as shown in Figure~\ref{fig:overview}.
Such irregular sample distribution prohibits us from many downstream applications described later. 

The resampling step aims at avoiding over- and undersampling by iteratively moving the base iso-points away from high-density regions, \ie
\begin{equation}\label{eq:repulsion_update}
\tilde{\q} \leftarrow \tilde{\q} - \alpha\vec{r},
\end{equation}
where $ \alpha = \sqrt{D/|\tilde{\Q}_t|}$ is the step size.
The update direction $ \vec{r} $ is a weighted average of the normalized translations between $\tilde{\q}$ and its K-nearest points (we set $K=8$):
\begin{equation}\label{eq:repulsion_direction}
	\vec{r} = \sum_{\tilde{\q}_i\in \N(\tilde{\q})}w(\tilde{\q}_i,\tilde{\q})\frac{\tilde{\q}_i - \tilde{\q}}{\|\tilde{\q}_i - \tilde{\q}\|} .
\end{equation} 
The weighting function is designed to gradually reduce the influence of faraway neighbors, specifically 
\begin{equation}\label{eq:spatial_w}
w(\tilde{\q}_i, \tilde{\q}) = e^{-\frac{\|\tilde{\q}_i-\tilde{\q}\|^2}{\sigma_p}},
\end{equation} 
where the density bandwidth $ \sigma_p $ is set to be $ 16D/|\tilde{\Q}_t| $.

\vspace{1mm}
\textbf{Upsampling.}
Next, we upsample the point set to the desired density while further improving the point distribution.
Our upsampling step is based on edge-aware resampling (EAR) proposed by Huang \etal\cite{huang2013edge}.
We explain the key steps and our main differences to EAR as follow.

First, we compute the normals as the normalized Jacobians and apply bilateral normal filtering, just as in EAR.
Then, the points are pushed away from the edges to create a clear separation. 
We modify the original optimization formulation with a simpler update consisting of an attraction and a repulsion term.
The former pulls points away from the edge and the latter prevents the points from clustering.
\begin{align}\label{eq:ear-push}
\Delta\p_{\text{attraction}} &= \dfrac{\sum_{\p_i\in \N(\p)}\phi(\n_i, \p_i - \p)(\p-\p_i)}{\sum_{\p_i\in \N(\p)}\phi(\n_i, \p_i - \p)}, \\
\Delta\p_{\text{repulsion}} &= 0.5\dfrac{\sum_{\p_i\in \N(\p)}w(\p_i,\q)(\p_i - \p)}{\sum_{\p_i\in \N(\p)}w(\p_i,\q)}, \\
\p & \leftarrow \p - \tau(\Delta\p_{\text{repulsion}}) - \tau(\Delta\p_{\text{attraction}}),
\end{align}
where $ \phi(\n_i, \p - \p_i) = e^{-\frac{(\n_i^\tr(\p - \p_i))^2}{\sigma_p}} $ is the anisotropic projection weight, $\n_i$ is the point normal of neighbor $ \p_i $ and $ w $ is the spatial weight defined in \eqref{eq:spatial_w}.
We use the same directional clipping function $ \tau $ as before to bound the two update terms individually, which improves the stability of the algorithm for sparse point clouds.

By design, new points are inserted in areas with low density or high curvature. 
The trade-off is controlled by a unifying priority score $ P(\p) = \max_{\p_i\in\N(\p)}B(\p, \p_i) $, where $ B $ is a distance measure (see \cite{huang2013edge} for the exact definition).
Denoting the point with the highest priority as $ \p^* $, a new point is inserted at the midpoint between $ \p^* $ and neighbor $ \p^*_{i^*} $, where $ \p^*_{i^*}=\arg\max_{\p_i\in\N(\p^*)} B(\p^*, \p_i) $.
In the original EAR method, the insertion is done iteratively, requiring an update of the neighborhood information and recalculation of $ B $ after every insertion.
Instead, to allow parallel computation at GPU, we approximate this process by simultaneously inserting a maximum of $ |\P|/10 $ points each step.
In this way, however, inserting the midpoint of point pairs would lead to duplicated new points.
Thus, we insert asymmetrically at $ \frac{1}{3}(\p^*_{i^*}+2\p^*) $ instead.

We then project the upsampled iso-points to the iso-surface. As shown in Figure~\ref{fig:overview}, the final iso-points successfully reflect the 3D geometry of the current implicit surface.

Compared with using marching cubes to extract the iso-surface, our adaptive sampling is efficient. Since both the resampling and upsampling steps only require information of local neighborhood, we implement it on the GPU.
Furthermore, since we use the iso-points from the previous iteration for initialization, the overall point distribution improves as the training stabilizes, requiring fewer (or even zero) resampling and upsampling steps at later stages.

\subsection{Utilizing iso-points in optimization}\label{sec:method-application}
We introduce two scenarios of using iso-points to guide neural implicit surface optimization: (i) importance sampling for multi-view reconstruction and (ii) regularization when reconstructing neural implicit surfaces from raw input point clouds.

\vspace{1mm}
\textbf{Iso-points for importance sampling.}
Optimizing a neural implicit function to represent a high-resolution 3D shape requires abundant training samples -- specifically, many supervision points sampled close to the iso-surface to capture the fine-grained geometry details accurately. 
However, for applications where the explicit 3D geometry is not available during training, the question of how to generate training samples remains mostly unexplored.

\begin{figure}[t!]
\centering
	\includegraphics[width=\linewidth]{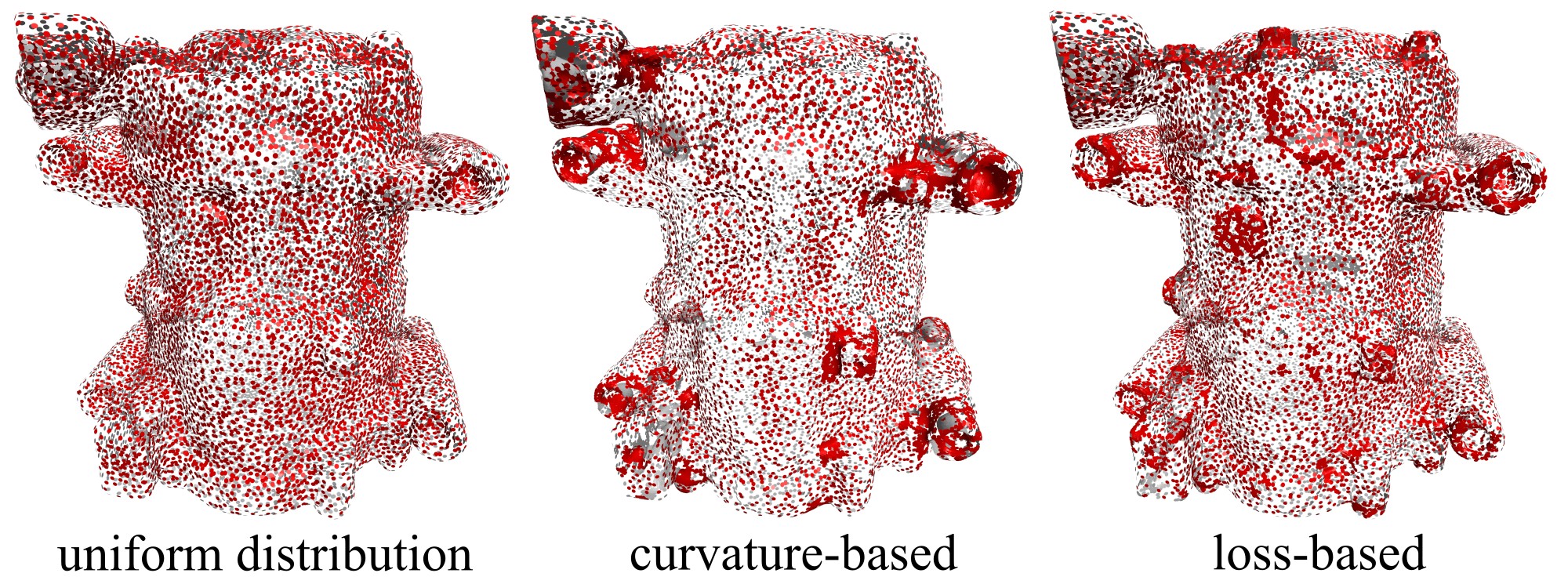}
	\caption{Examples of importance sampling based on different saliency metrics. The default uniform iso-points treat different regions of the iso-surface equally. We compute different saliency metrics on the uniform iso-points, based on which more samples can be gathered in the salient regions. The curvature-based metric emphasizes geometric details, while the loss-based metric allows for hard example mining.}\label{fig:metric_sampling}
\end{figure}

We exploit the geometry information and the prediction uncertainty carried by the iso-points during training. 
The main idea is to compute a saliency metric on iso-points, then add more samples in those regions with high saliency.
To this end, we experiment with two types of metrics: curvature-based and loss-based.
The former aims at emphasizing geometric features, typically reflected by high curvature. The latter is a form of hard example mining, as we sample more densely where the higher loss occurs, as shown in Figure~\ref{fig:metric_sampling}.

Since the iso-points are uniformly distributed, the curvature can be approximated by the norm of the Laplacian, \ie $ \mathcal{R}_{\text{curvature}}(\p) = \|\p-\sum_{\p_i\in\N(\p)} \p_i\| $.
For the loss-based metric, we project the iso-points on all training views and compute the average loss at each point, \ie $ \mathcal{R}_{\text{loss}}(\p) = \frac{1}{N}\sum_{i}^{N} \text{loss}(\p)$, where $ N $ is the number of occurrences of $ \p $ in all views.

As both metrics evolve smoothly, we need not update them in each training iteration.
Denoting the iso-points at which we computed the metric as $ \mathcal{T} $ and the subset of template points with high metric values by $ \mathcal{T}^*=\{\vec{\vec{t}^*}\} $, the metric-based insertion for each point $ \p $ in the current iso-point set $ \P_t $ can be written as
\begin{equation*}\small
\textstyle
\p_{\text{new},i} = \frac{2}{3}\p + \frac{1}{3}\p_i, \forall \p_i \in \N(\p), \;\text{if } \min_{\vec{t}^*}\|\p-\vec{t}^*\|\leq \sigma.
\end{equation*}
The neighborhood radius $ \sigma $ is the same one used in \eqref{eq:repulsion_direction}.

\vspace{1mm}
\textbf{Iso-points for regularization.}
The access to an explicit representation of the implicit surface also enables us to incorporate geometry-motivated priors into the optimization objective, exerting finer control of the reconstruction result.

\begin{figure}[t!]
\centering
	\includegraphics[width=0.3\linewidth]{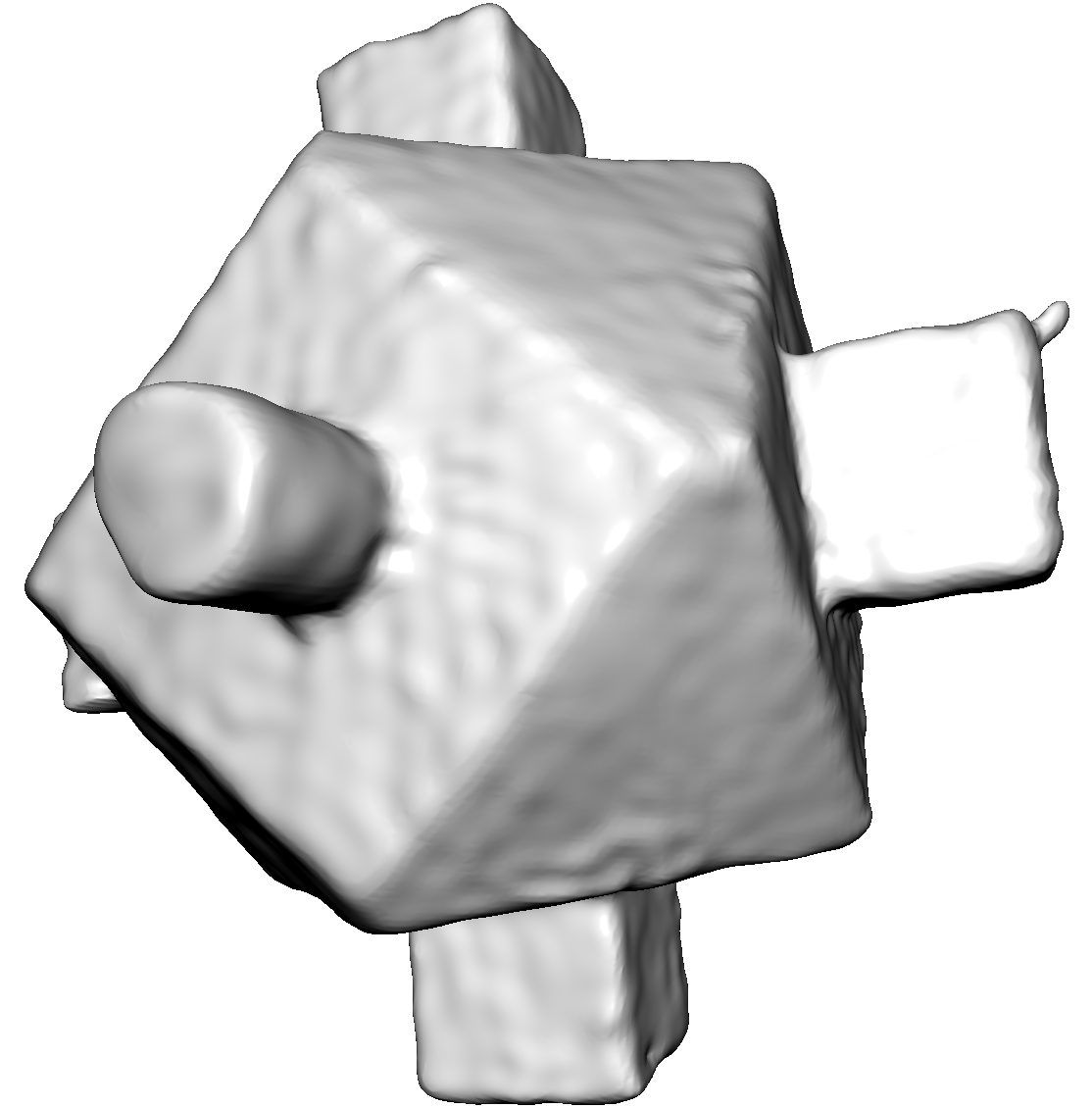}
	\includegraphics[width=0.3\linewidth]{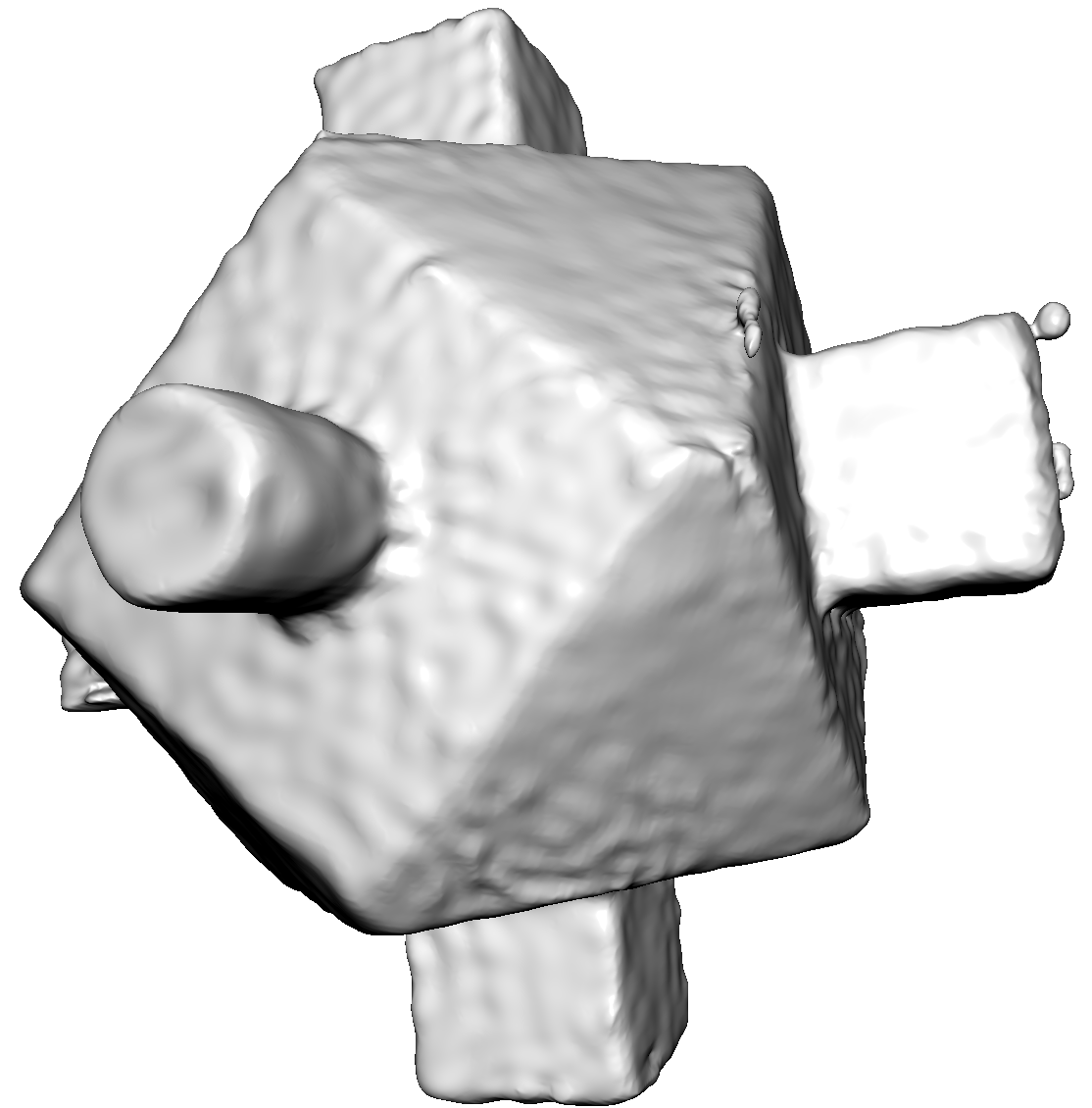}
	\includegraphics[width=0.3\linewidth]{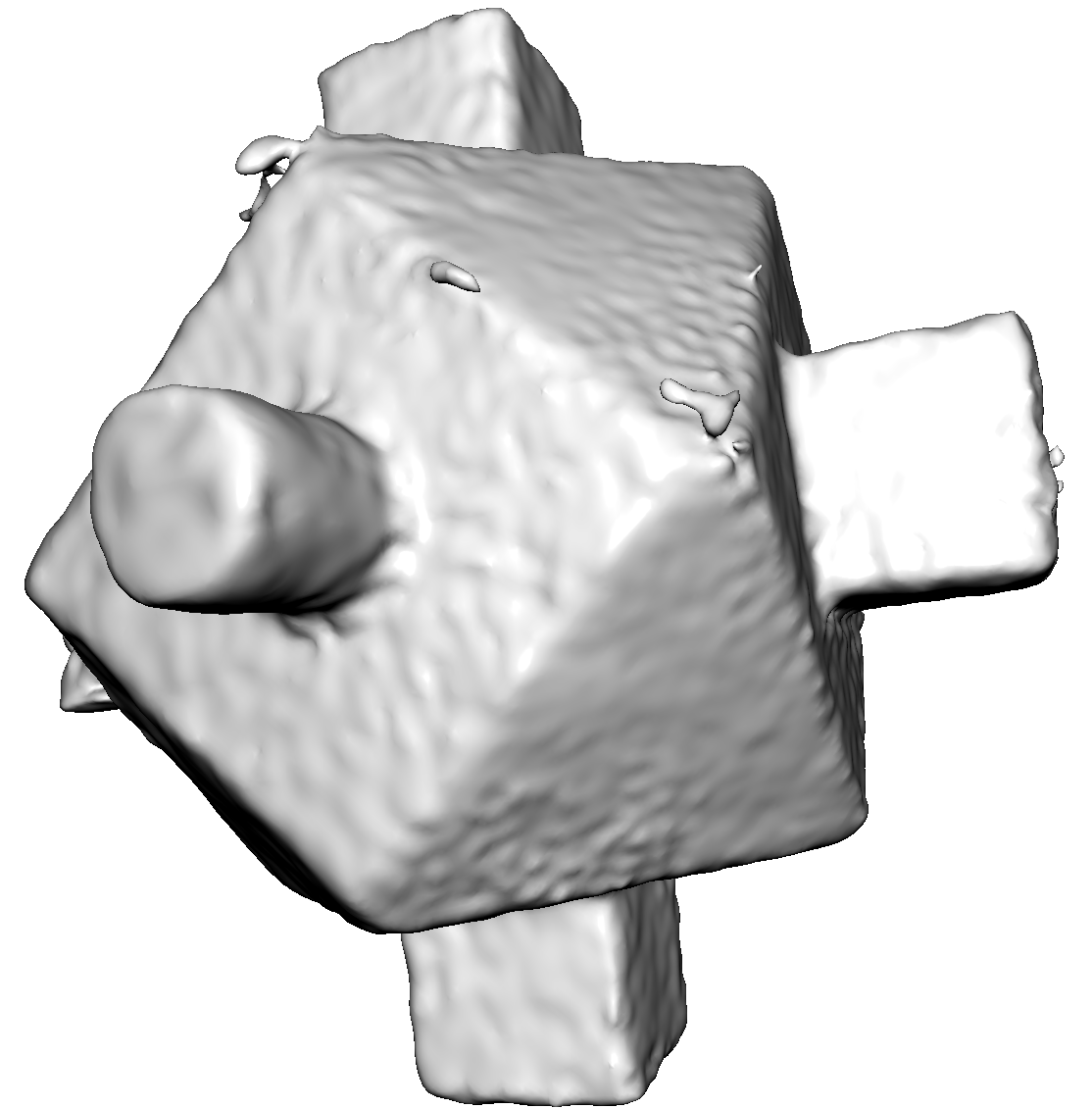}
\caption{Progression of overfitting. When optimizing a neural implicit surface on a noisy point cloud, the network initially outputs a smooth surface, but increasingly overfits to the noise in the data. Shown here are the reconstructed surfaces after 1000, 2000 and 5000 iterations. The input point cloud is acquired in-house using an Artec Eva scanner.}\label{fig:overfitting}
\end{figure}

Let us consider fitting a neural implicit surface to a point cloud.
Depending on the acquisition process, the point cloud may be sparse, distorted by noise, or distributed highly unevenly.
Similar to previous works \cite{ulyanov2018deep, williams2019deep}, we observe that the neural network tends to reconstruct a smooth surface at the early stage of learning, but then starts to pick up higher frequency signals from the corrupted observations and overfits to noise and other caveats in the data,  as shown in Figure~\ref{fig:overfitting}.
This characteristic is consistent across network architectures, including those designed to accommodate high-frequency information, such as \textsc{SIREN}.

Existing methods that address overfitting include early stopping, weight decay, drop-out \etc\cite{goodfellow2016deep}.
However, whereas these methods are generic tools designed to improve network generalizability for training on large datasets, we propose a novel regularization approach that is specifically tailored to training neural implicit surfaces and serves as a platform to integrate a plethora of 3D priors from classic geometry processing.

Our main idea is to use the iso-points as a consistent, smooth, yet dynamic approximation of the reference geometry.
Consistency and smoothness ensure that the optimization does not fluctuate and overfit to noise; the dynamic nature lets the network pick up consistent high-frequency signals governed by underlying geometric details.

To this end, we extract iso-points after a short warmup training (\eg 300 iterations).
Because of the aforementioned smooth characteristic of the network, the noise level in the initial iso-points is minimal.
Then, during subsequent training, we update the iso-points periodically (\eg every 1000 iterations) to allow them to gradually evolve as the network learns more high-frequency information.

The utility of the iso-points includes, but is not limited to
\begin{inparaenum}[1)]
\item serving as additional training data to circumvent data scarcity, 
\item enforcing additional geometric constraints,
\item filtering outliers in the training data.
\end{inparaenum}

Specifically, for sparse or hole-ridden input point clouds, we take advantage of the uniform distribution of iso-points and augment supervision in undersampled areas by enforcing the signed distance value on all iso-points to be zero:
\begin{equation}\label{eq:iso_sdf}
\mathcal{L}_{\text{isoSDF}} = \dfrac{1}{|\P|}\sum_{\p\in \P}|f(\p)|.
\end{equation}

Given the iso-points, we compute their normals from their local neighborhood using principal component analysis (PCA)~\cite{hoppe1992surface}. 
We then increase surface smoothness by enforcing consistency between the normals estimated by PCA and those computed from the gradient of the network, \ie
\begin{equation}\label{eq:iso_normal}
\mathcal{L}_{\text{isoNormal}} = \dfrac{1}{|\P|}\sum_{\p\in \P}(1-|\cos(J_f^\tr(\p), \n_{\text{PCA}})|).
\end{equation}
The larger the PCA neighborhood is, the smoother the reconstruction becomes.
Optionally, additional normal filtering can be applied after PCA to reduce over-smoothing and enhance geometric features.

\begin{figure}[t!]
\centering
\includegraphics[height=0.3\linewidth]{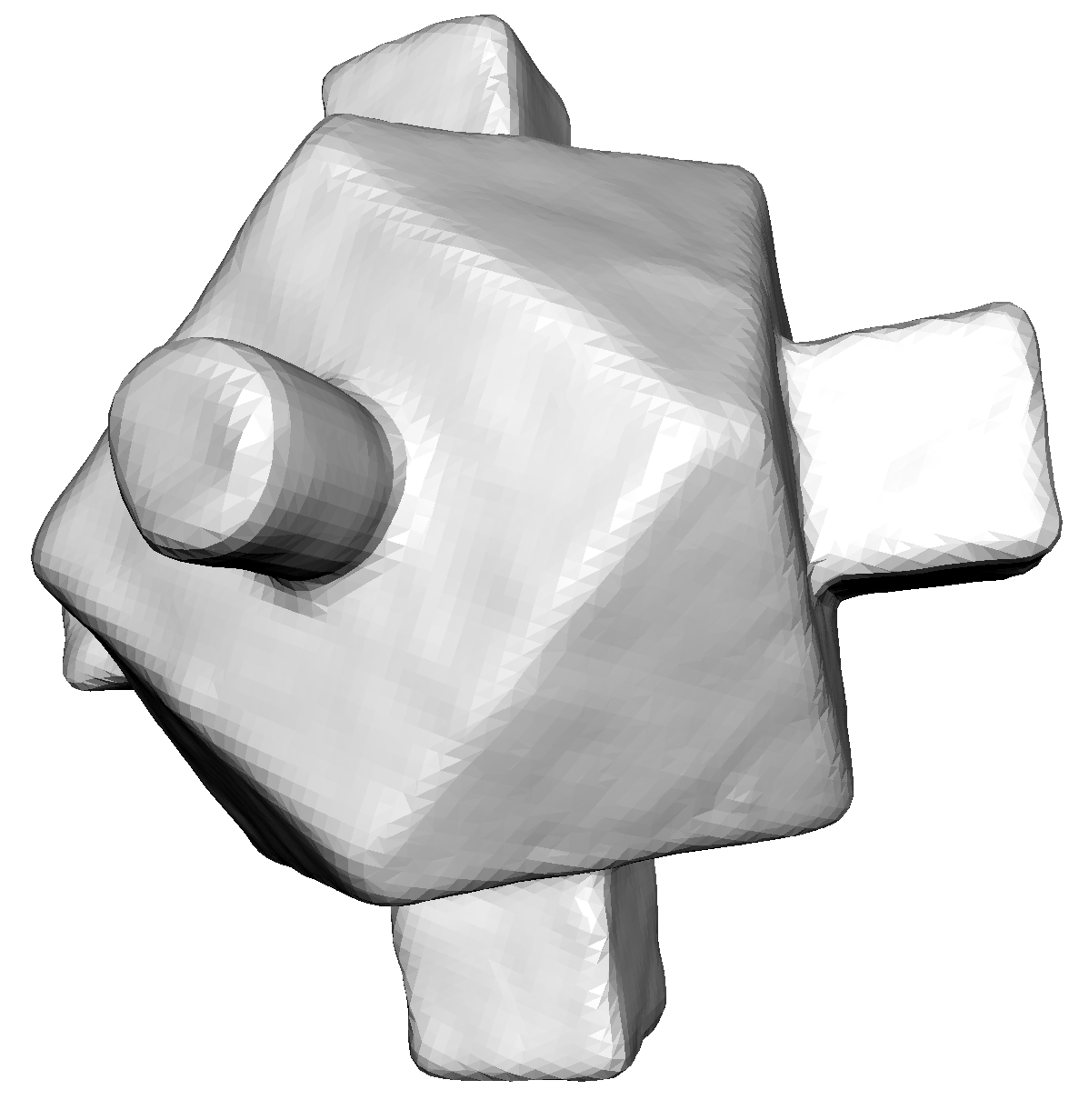}
\includegraphics[height=0.3\linewidth,trim={4cm 0 0 0},clip]{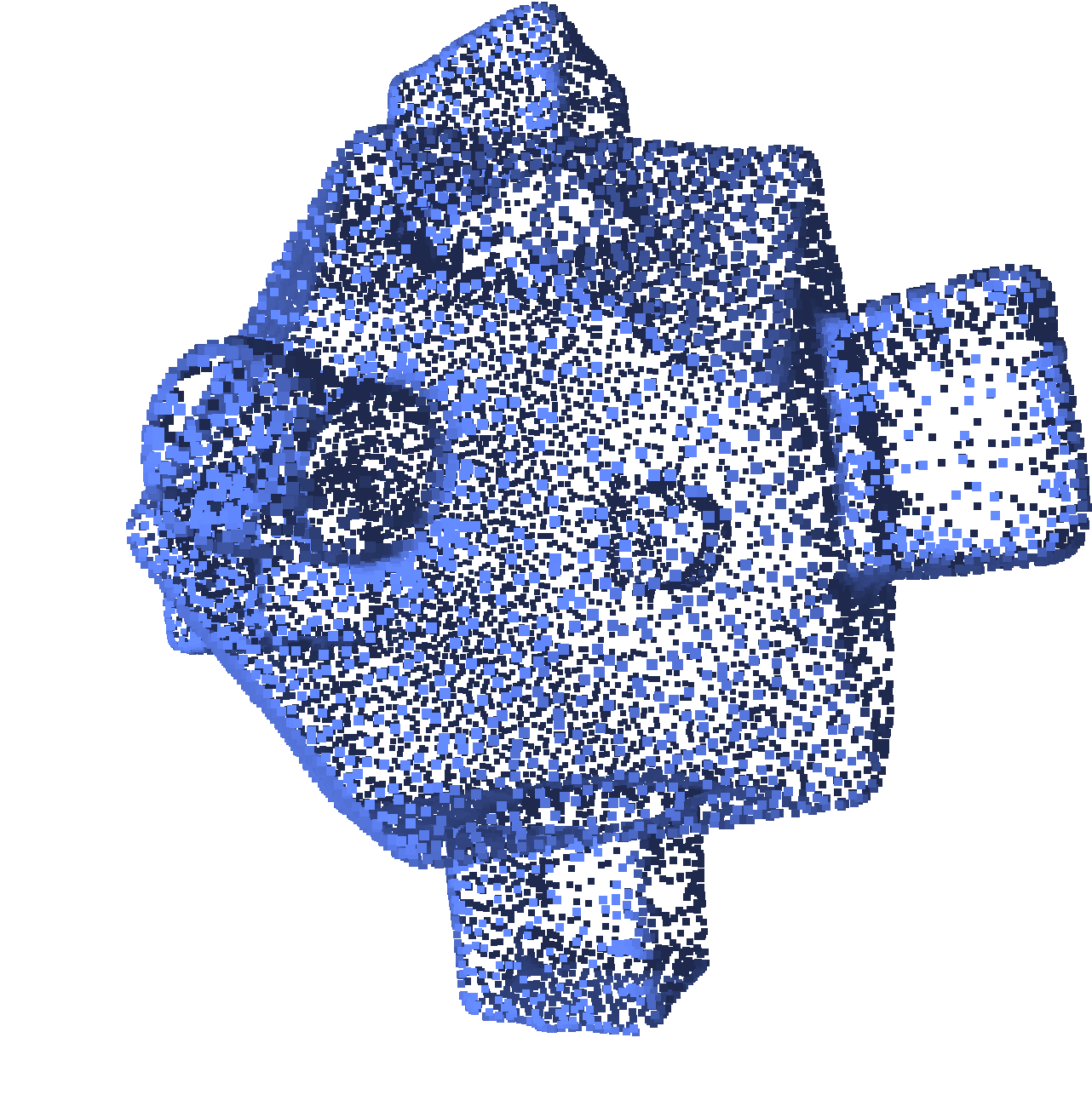}
\includegraphics[height=0.3\linewidth]{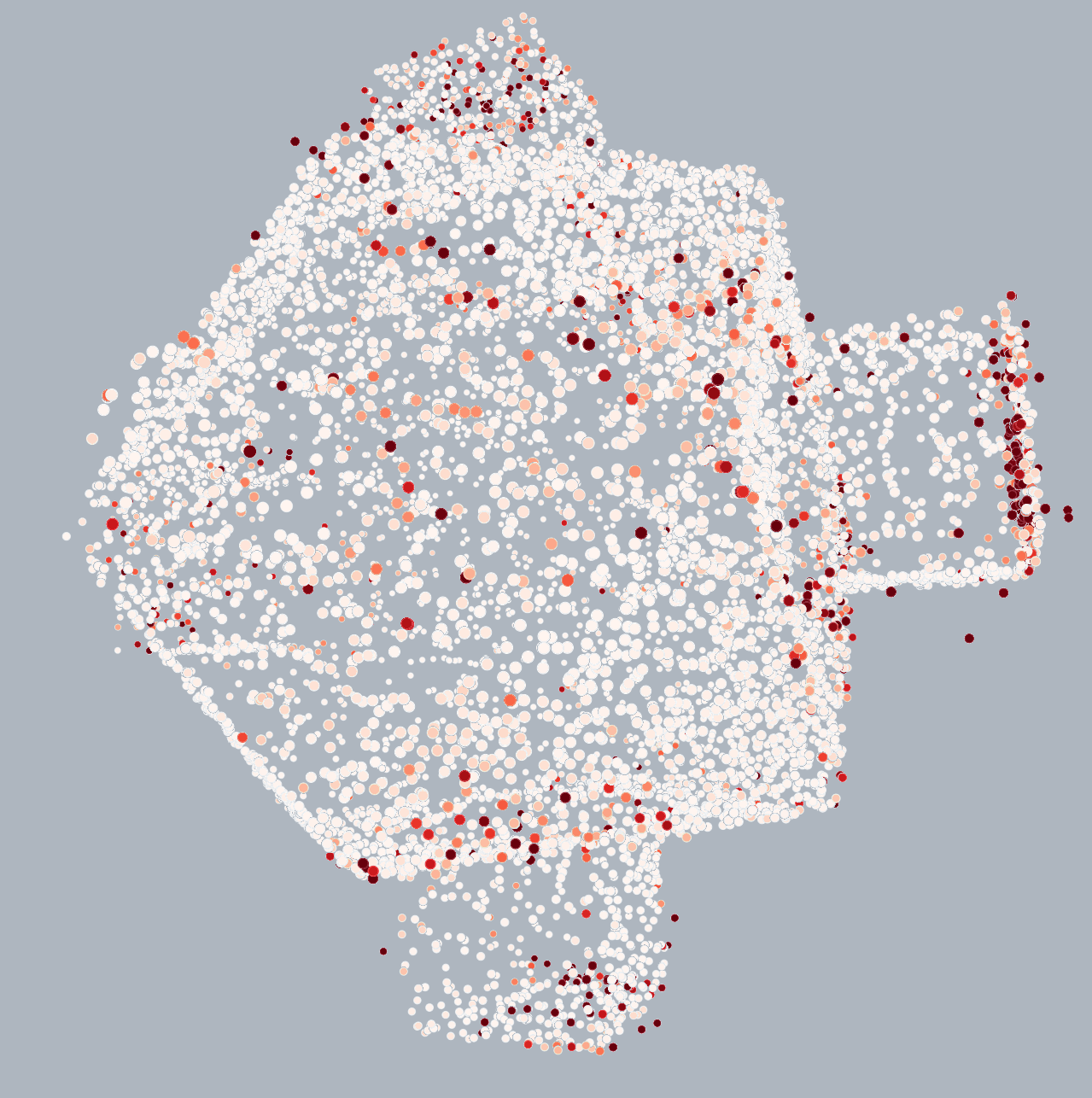}
\caption{Iso-points for regularization. At the early stage of the training, the implicit surface is smooth (left), and we extract iso-points (middle) as a reference shape, which can facilitate various regularization terms. In the example on the right, we use the iso-points to reduce the influence of outliers (shown in red). }\label{fig:reg}
\end{figure}

Finally, we can use the iso-points to filter outliers in the training data.
Specifically, given a batch of training points $ \Q=\{\q\} $, we compute a per-point loss weight based on their alignment with the iso-points.
Here, we choose to use bilateral weights to take both the Euclidean and the directional distance into consideration.
Denoting the normalized gradient of an iso-point $ \p $ and a training point $ \q $ as $ \n_\p $ and $ \n_\q $, respectively, the bilateral weight can written as
\begin{gather}
v(\q)  = \min_{\p\in\P}\phi(\n_\p, \p-\q)\psi(\n_\p, \n_\q),\quad \text{with}\\\label{eq:outlier_weight}
\psi(\n_\p, \n_\q) =e^{-\left(1-\frac{1-\n_\p^\tr\n_\q}{1-\cos\left(\sigma_n\right)}\right)^2},
\end{gather}
where $ \sigma_n $ regulates the sensitivity to normal difference; we set $ \sigma_n = 60^\circ $ in our experiments.
This loss weight can be incorporated into the existing loss functions to reduce the impact of outliers.
A visualization of the outliers detected by this weight is shown in Figure~\ref{fig:reg}.

\begin{figure}
	\includegraphics[width=\linewidth]{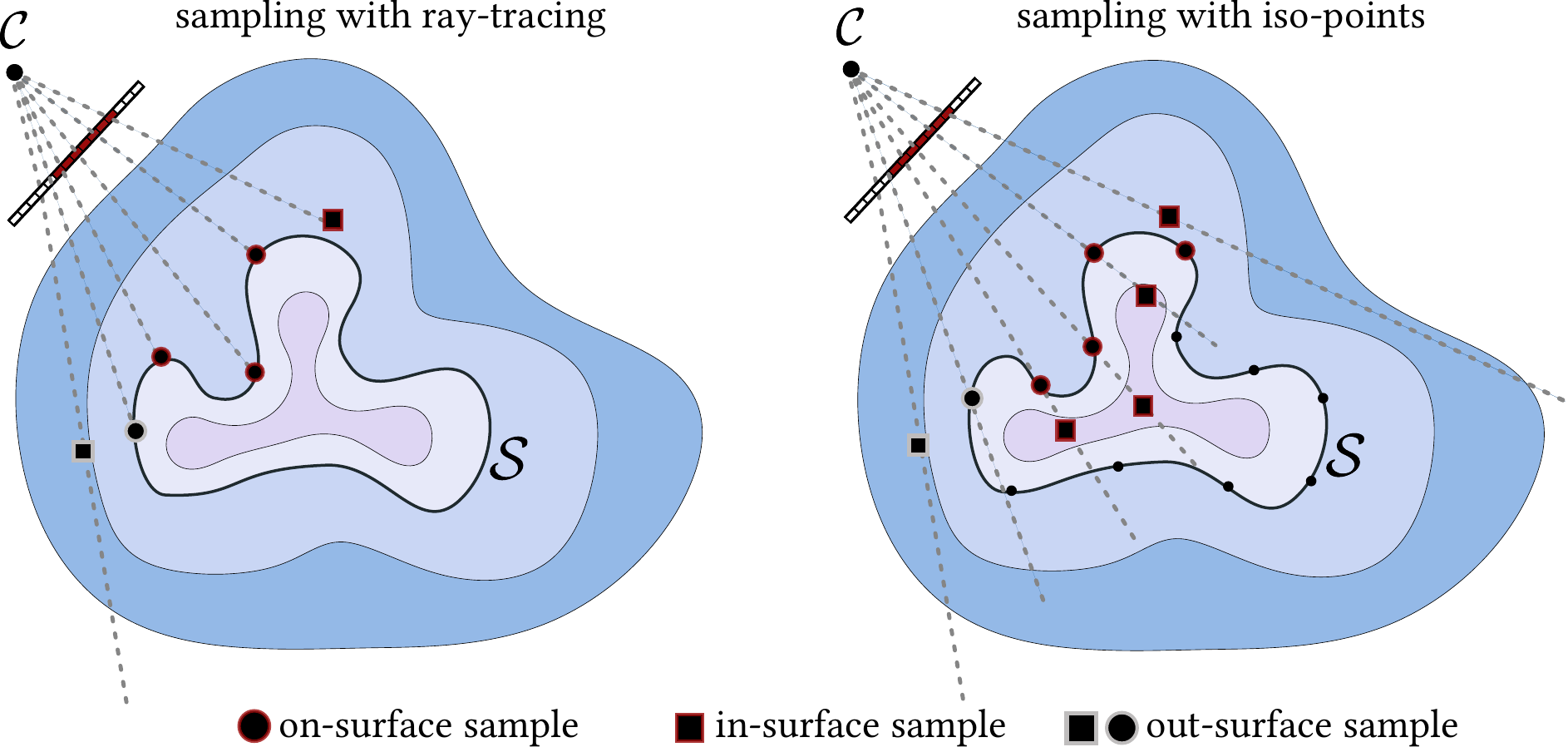}
	\caption{A 2D illustration of two sampling strategies for multi-view reconstruction. Ray-tracing (left) generates training samples by shooting rays from camera $ \mathcal{C} $ through randomly sampled pixels; depending on whether an intersection is found and whether the pixel lies inside the object silhouette, three types of samples are generated: on-surface, in-surface and out-surface points. We generate on-surface samples directly from iso-points, obtaining evenly distributed samples on the implicit surface, and also use the iso-points to generate more reliable in-surface samples. }\label{fig:ray-march}
\end{figure}

\section{Results}\label{sec:result}
Iso-points can be incorporated into the optimization pipelines of existing state-of-the-art methods for implicit surface reconstruction.
In this section, we demonstrate the benefits of the specific techniques introduced in Section~\ref{sec:method-application}.

We choose state-of-the-art methods as the baselines, then augment the optimization with iso-points.
Results show that the augmented optimization outperforms the baseline quantitatively and qualitatively.

\subsection{Sampling with iso-points}
We evaluate the benefit of utilizing iso-points to generate training samples for multi-view reconstruction.
As the baseline, we employ the ray-tracing algorithm from a state-of-the-art neural implicit renderer IDR \cite{yariv2020multiview}, which generates training samples by ray-marching from the camera center through uniformly sampled pixels in the image.
As shown in Figure~\ref{fig:ray-march}, three types of samples are used for different types of supervision: on-surface samples, which are ray-surface intersections inside the object's 2D silhouette, in-surface samples, which are points with the lowest signed distance on the non-intersecting rays inside 2D silhouette, and out-surface samples, which are on the rays outside the 2D silhouette either at the surface intersection or at the position with the lowest signed distance. 

On this basis, we incorporate the iso-points directly as on-surface samples. We can direct the learning attention by varying the distribution of iso-points using the saliency metrics described in Section~\ref{sec:method-application}. The iso-points also provide us prior knowledge to generate more reliable in-surface samples.
More specifically, as shown in Figure~\ref{fig:ray-march} (right), we generate the three types of samples as follows:
\begin{inparaenum}[a)]
\item on-surface samples: we remove occluded iso-points by visibility testing using a point rasterizer, and select those iso-points whose projections are inside the object silhouette;
\item in-surface samples: on the camera rays that pass through the on-surface samples, we determine the point with the lowest signed distance on the segment between the on-surface sample and the farther intersection with the object's bounding sphere.
\item out-surface samples: we shoot camera rays through pixels outside the object silhouette, and choose the point with the lowest signed distance.
\end{inparaenum}
\begin{table}
	\centering\scriptsize
	\begin{tabular}{lcccc}
		\toprule
		& \makecell[t]{baseline\\(ray-tracing)} & \makecell[t]{uniform} & \makecell[t]{curvature-\\based} & \makecell[t]{loss-\\based} \\\midrule
	\makecell[l]{CD $ \cdot 10^{-4} $(position)} & 17.24 & 1.80	& 1.83	& \textbf{1.71} \\
	\makecell[l]{CD $ \cdot 10^{-1} $(normal)}  & 1.51 & 1.10 & 0.99 & \textbf{0.95}\\\bottomrule
	\end{tabular}
	\caption{Quantitative effect of importance sampling with iso-points. Compared to the baseline, which generates training points via ray-marching, we use iso-points to draw more attention on the implicit surface. The result is averaged over 10 models selected from the Sketchfab dataset \cite{yifan2019patch}.}\label{tab:sampling}
\end{table}%

Below, we demonstrate two benefits of the proposed sampling scheme.
\begin{figure}
\centering
{\includegraphics[width=.95\linewidth]{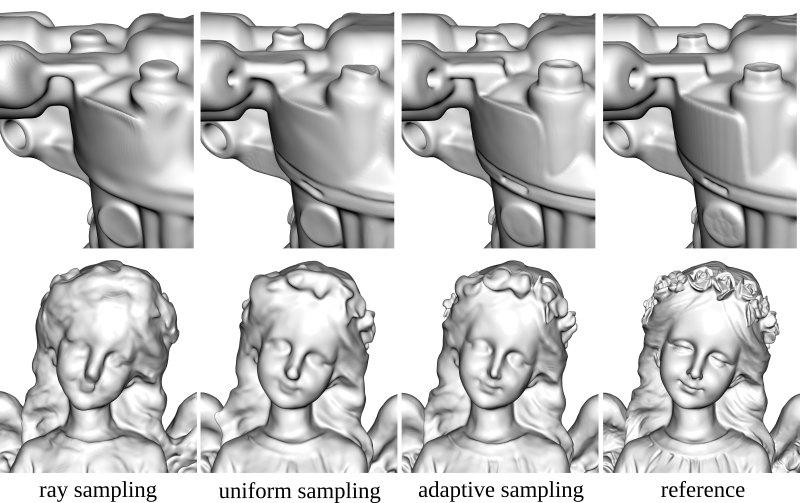}}
\caption{Qualitative comparison between sampling strategies for multi-view reconstruction. Using the same optimization time and similar total sample count, sampling the surface points with uniformly distributed iso-points considerably improves the reconstruction accuracy. A loss based importance sampling further improves the recovery of small-scale structures. The models shown here are the \textsc{compressor} and \textsc{angel2} from the Sketchfab dataset \cite{yifan2019patch}.}
\label{fig:compressor}
\end{figure}

\vspace{1mm}
\textbf{Surface details from importance sampling.} 
First, we examine the effect of drawing on-surface samples using iso-points by comparing the optimization results under fixed optimization time and the same total sample count.

As inputs, we render $ 512\times512 $ images per object under known lighting and material from varied camera positions.
When training with the iso-points, we extract 4,000 iso-points after 500 iterations, then gradually increase the density until reaching 20,000 points.
To match the sample count, in the ray-tracing variation, we randomly draw 2,048 pixels per image and then increase the sample count until reaching 10,000 pixels.
We use a 3-layer \textsc{SIREN} model with the frequency multipliers set to 30, and optimize with a batch size of 4.

We evaluate our method quantitatively using 10 watertight models from the Sketchfab dataset \cite{yifan2019patch}. 
As shown in  Table~\ref{tab:sampling}, 
we compute 2-way chamfer point-to-point distance ($ \|\p_i - \p_j\|^2 $) and normal distance \hbox{$(1- \cos(\vec{n}_i, \vec{n}_j) )$} on 50K points, uniformly sampled from the reconstructed meshes.

Results show that using uniform iso-points as on-surface samples compares favorably against the baseline, especially in the normal metric. It suggests that we achieve higher fidelity on the finer geometric features, as our surface samples overcome under-sampling issues occurring at small scale details.
We also see that importance sampling with iso-points and loss based upsampling exhibits a substantial advantage over other variations, demonstrating the effectiveness of smart allocation of the training samples according to the current learning state.
In comparison, curvature-based sampling performs similarly to the baseline, but notability worse than with the uniform iso-points.
We observe that the iso-points, in this case, are highly concentrated on a few spots on the surface and ignore regions where the current reconstruction is problematic (Figure~\ref{fig:metric_sampling}).

The improvement is more pronounced qualitatively, as shown in Figure~\ref{fig:compressor}.
Sampling on-surface with uniform iso-points clearly enhances reconstruction accuracy compared to the baseline with ray-tracing.
The finer geometric details further improve with loss-based importance sampling.

\begin{figure}
	\includegraphics[width=0.95\linewidth]{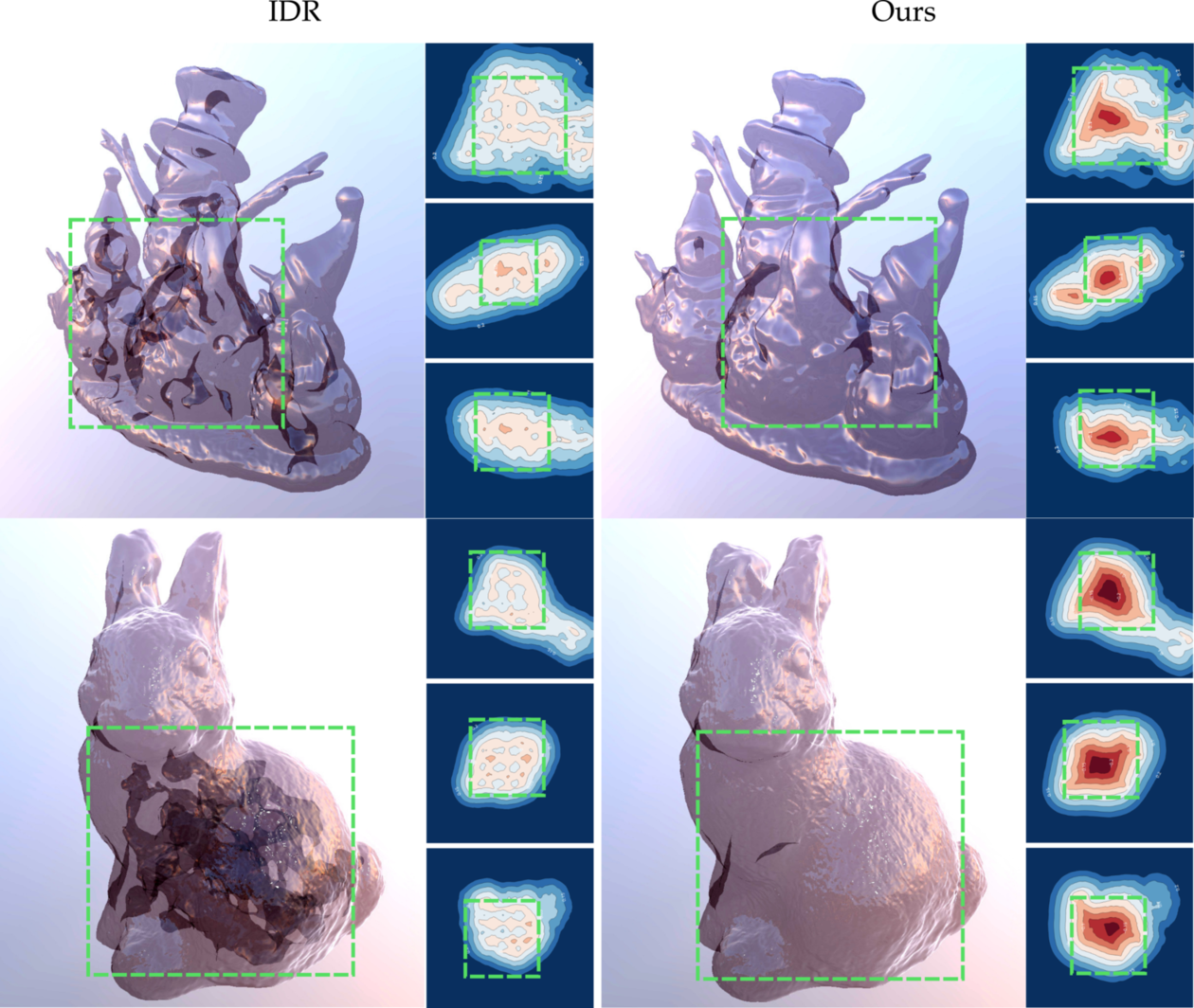}
	\caption{Topological correctness of the reconstructed surface in multi-view reconstruction. The erroneous inner structures of IDR are indicated by the artifacts in the images rendered by Blender~\cite{Hess:2010:BFE:1893021} (using glossy and transparent material) and by the contours of the signed distance field on cross-sections.}\label{fig:dtu}
\end{figure}

\vspace{1mm}
\textbf{Topological correctness from 3D prior.}
IDR can reconstruct impressive geometric details on the DTU dataset~\cite{jensen2014large}, but a closer inspection shows that the reconstructed surface contains a considerable amount of topological errors inside the visible surface.
We use iso-points to improve the topological accuracy of the reconstruction.

We use the same network architecture and training protocol as IDR, which samples 2048 pixels from a randomly chosen view in each optimization iteration.
We use uniform iso-points in this experiment.
To keep a comparable sample count, we subsample the visible iso-points to obtain a maximum of 1500 on-surface samples per iteration.
Since our strategy automatically creates more in-surface samples (as shown in Figure~\ref{fig:ray-march}), we halve the loss weight on the in-surface samples compared to the original implementation.

We visualize the topology of the reconstructed surface in Figure~\ref{fig:dtu}.
To show the inner structures of the surface, we render it in transparent and glossy material with a physically-based renderer~\cite{Hess:2010:BFE:1893021} and show the back faces of the mesh.
Dark patches in the rendered images indicate potentially erroneous light transmission caused by inner structures. 
Similarly, we also show the contour lines of the iso-surface on a cross-section to indicate the irregularity of the reconstructed implicit function.
In both visualizations, the incorrect topology in IDR reconstruction is apparent. 
In contrast, our sampling enables more accurate reconstruction of the signed distance field inside and outside the surface with more faithful topological structure.


\subsection{Regularization with iso-points}
We evaluate the benefit of using iso-points for regularization.
As an example application, we consider surface reconstruction from a noisy point cloud.

As our baseline method, we use the publicly available \textsc{SIREN} codebase and adopt their default optimization protocol.
The noisy input point clouds are either acquired with a 3D scanner (Artec Eva) or reconstructed~\cite{furukawa2009accurate} from the multi-view images in the DTU dataset.

In each optimization iteration, the baseline method randomly samples an equal number of oriented surface points $ \Q_s = \{\q_s, \n_s\} $ from the input point cloud and unoriented off-surface points $ \Q_o = \{\q_o\} $ from bounding cube's interior.
The optimization objective is comprised of four parts:
\begin{center}
  \scalebox{0.85}{%
  $\displaystyle 
  \mathcal{L} = \gamma_{\text{onSDF}}\mathcal{L}_{\text{onSDF}} +  \gamma_{\text{normal}}\mathcal{L}_{\text{normal}} +  
		\gamma_{\text{offSDF}}\mathcal{L}_{\text{offSDF}}+\gamma_{\text{eikonal}} \mathcal{L}_{\text{eikonal}},
  $}
\end{center}
where
\begin{center}
\scalebox{0.8} {
$	\mathcal{L}_{\text{onSDF}} = \frac{\sum_{\q_s\in \Q_s}|f(\q_s)|}{|\Q_s|}, \, \mathcal{L}_{\text{normal}} = \frac{\sum_{\q_s\in \Q_s}|1-\cos(J_f^\tr(\q_s), \n_s)|}{|\Q_s|}, 
$
}
\end{center}

\vspace{-3mm}

\begin{center}
\scalebox{0.8} {
$ 
	\mathcal{L}_{\text{offSDF}} = \frac{\sum_{\q_o\in \Q_o}e^{-\alpha|f(\q_o)|}}{|\Q_o|}, \, \mathcal{L}_{\text{eikonal}} = \frac{\sum_{\q\in \Q_o\cup\Q_s }|1-\|J_f^\tr(\q)\||}{|\Q_s|+|\P_o|}
$ and
}
\end{center}
\[\textstyle {\gamma_{\text{onSDF}} = 1000, \gamma_{\text{normal}} = 100,  \gamma_{\text{offSDF}} = 50, \gamma_{\text{eikonal}}=100}. \]
We alter this objective with the outlier-aware loss weight defined in \eqref{eq:outlier_weight}, and then add the regularizations on iso-points $ \mathcal{L}_\text{isoSDF} $ \eqref{eq:iso_sdf} and $ \mathcal{L}_{\text{isoNormal}} $ \eqref{eq:iso_normal}. 
The final objective becomes 

\begin{center}
\scalebox{0.8} {
$ 
	\mathcal{L} = \gamma_{\text{onSDF}}(\mathcal{L}_{\text{onSDF}}+\mathcal{L}_{\text{isoSDF}}) +  \gamma_{\text{normal}}(\mathcal{L}_{\text{normal}}+\mathcal{L}_{\text{isoNormal}}) + $
}
\end{center}
\vspace{-6mm} 
\begin{center}
\scalebox{0.79} {
$ 
		\gamma_{\text{offSDF}}\mathcal{L}_{\text{offSDF}}+\gamma_{\text{eikonal}} \mathcal{L}_{\text{eikonal}},$
}
\end{center}
where the loss terms with on-surface points are weighted as follows:

\noindent
\scalebox{0.9} {
$ 
	\phantom{aaaaa}\mathcal{L}_{\text{onSDF}} = \frac{1}{|\Q_s|}{\sum_{\q_s\in \Q_s}v(\q_s)|f(\q_s)|}, 
	$
}
\scalebox{0.9} {
$ 
	\phantom{aaaaa}\mathcal{L}_{\text{normal}} = \frac{1}{{|\Q_s|}}{\sum_{\q_s\in \Q_s}v(\q_s)|1-\cos(J_f^\tr(\q_s), \n_s)|}.
	$
}

\vspace{2mm}
\noindent The iso-points are initialized by subsampling the input point cloud by 1/8 and updated every 2000 iterations.

\begin{figure}
\includegraphics[width=\linewidth]{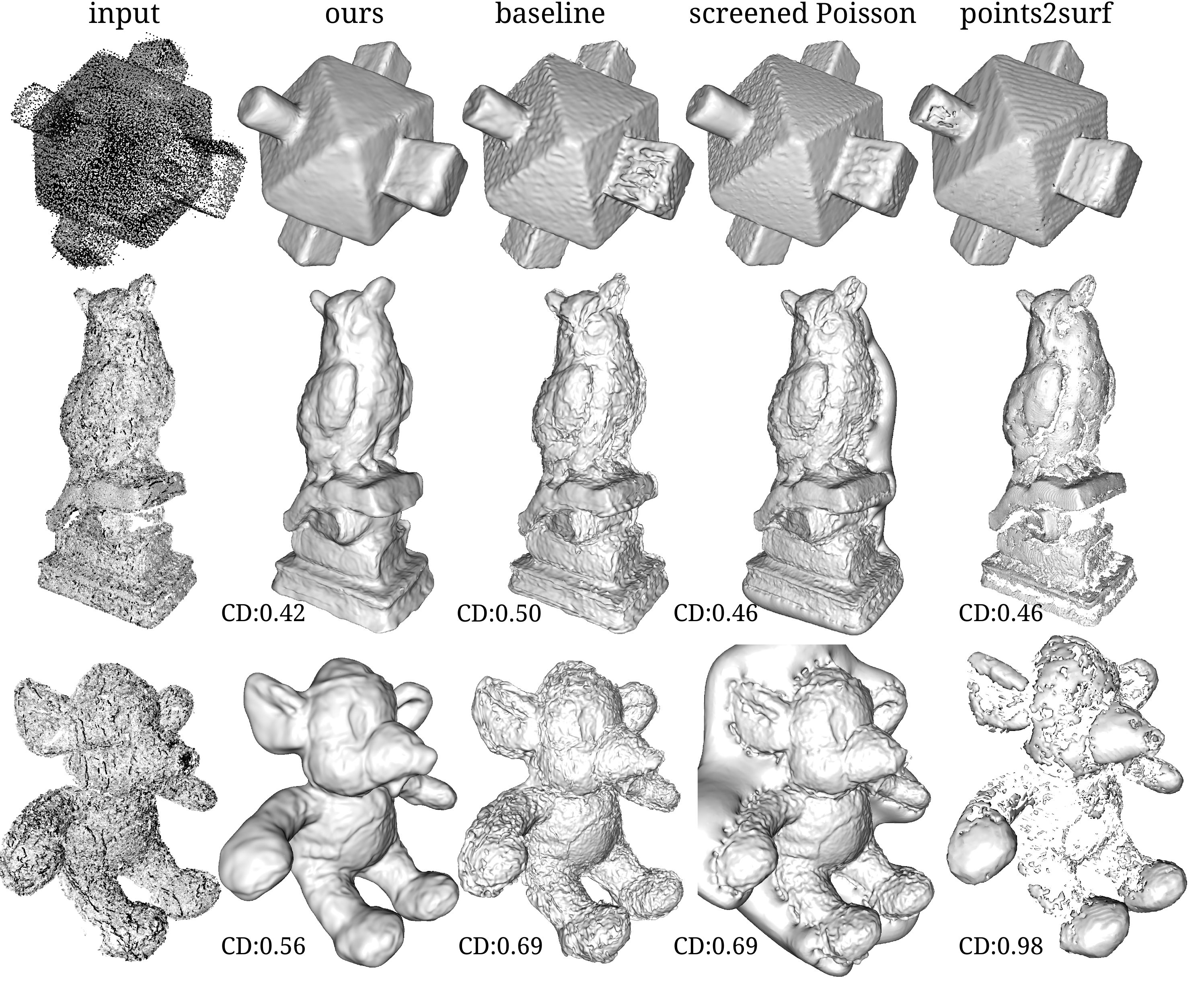}
\caption{Implicit surface reconstruction from noisy and sparse point clouds. From left to right: input, reconstruction with our proposed regularizations, baseline reconstruction without our regularizations, screened Poisson reconstruction and Points2Surf. CD denotes L1-Chamfer distance. The sparse point cloud in the first row is acquired with an Artec Eva scanner, whereas the inputs in the second row and third row are reconstructed from DTU dataset (model 105 and 122) using \cite{furukawa2009accurate}.}\label{fig:owl}
\end{figure}

The comparison with the baseline, i.e., vanilla optimization without regularization, is shown in Figure~\ref{fig:owl}. 
For the DTU-MVS data, we also conduct quantitative evaluation following the standard DTU protocol as shown in Table~\ref{tab:pclinputs}, i.e., L1-Chamfer distance between the reconstructed and reference point cloud within a predefined volumetric mask~\cite{aanaes2016large}.
For clarity, we also show the results of screened Poisson reconstruction~\cite{kazhdan2013screened} and Points2Surf~\cite{erler2020points2surf}.
The former reconstructs a watertight surface from an oriented point set by solving local Poisson equations; the latter fits an implicit neural function to an unoriented point set in a global-to-local manner.
Compared with the baseline and screened Poisson, our proposed regularizations significantly suppresses noise.
Points2Surf can handle noisy input well, but the sign propagation appears to be sensitive to the point distribution, leading to holes in the reconstructed mesh.
Moreover, since their model does not use the points' normal information, the reconstruction lacks detail.

\begin{table}
\centering\scriptsize
\begin{tabular}{ccccc}
	\toprule
	ID	& ours & baseline & point2surf & screened Poisson \\\midrule
	55  & \textbf{0.37} &  0.41 & 0.56 & 0.42  \\
	69  & \textbf{0.59} & 0.65 & 0.61 & 0.63  \\
	105 & \textbf{0.56}  & 0.69 & 0.98 & 0.69  \\
	110  & 0.54 & \textbf{0.51} & 0.61 & 0.55 \\
	114  & \textbf{0.38}& 0.45 & 0.45 & 0.37 \\
	118  & \textbf{0.45}& 0.49 & 0.59 & 0.55  \\
	122  & \textbf{0.42}& 0.50 & 0.46 & 0.46  \\
	Average & 0.53 & 0.61 & 0.52 & \textbf{0.47} 
	\\\bottomrule
\end{tabular}
\caption{Quantitative evaluation for surface reconstruction from a noisy sparse point cloud. We evaluate the two-way L1-chamfer distance on a subset of the DTU-MVS dataset. }\label{tab:pclinputs}
\end{table}
\subsection{Performance analysis}
The main overhead in our approach is the projection step. One newton iteration requires a forward and a backward pass.
On average, the projection terminates within 4 iterations. 
This procedure is optimized by only considering points that are not yet converged at each iteration.

Empirically, the computation time of extracting the iso-points once is typically equivalent to running 3 training iterations.
In practice, as we extract iso-points only periodically, the total optimization time only increases marginally.

In the case of multi-view reconstruction, since the ray-marching itself is an expensive operation, involving multiple forward passes per ray, the overhead of our approach is much less notable.
As discussed, we filter the occluded iso-points before the projection, which also saves optimization time.
The trade-off between optimization speed
\setlength{\intextsep}{4pt}%
\setlength{\columnsep}{4mm}%
\begin{wrapfigure}{r}{0.60\linewidth}\scriptsize
	\begin{tikzpicture}
		\begin{semilogyaxis}[
			clip=true,
			height=4.5cm, width=\linewidth,
			xlabel=time (s),
			ylabel=chamfer distance]
			\addplot[color=red,mark=x] 
			table[x=mtime,y=chamfer_p,col sep=comma,comment chars=\#] {data/compressor_implicit.log};		
			\addplot[color=blue,mark=x] 
			table[x=mtime,y=chamfer_p,col sep=comma,comment chars=\#] {data/compressor_lossS.log};
			\legend{ray-marching, iso-points}
		\end{semilogyaxis}
	\end{tikzpicture}\vspace{-3mm}
	\caption{Validation error in relation to optimization time. The first time stamp is at the \hbox{$ 100 $-th} iteration. }\label{fig:training_time}	
\end{wrapfigure}
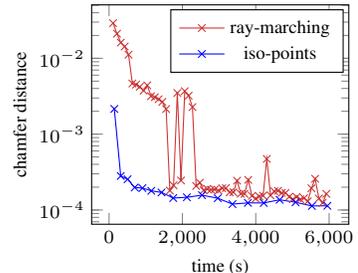
  and quality is depicted in a concrete example in Fig.~\ref{fig:training_time}, where we plot the evolution of the chamfer distance during the optimization of the \textsc{compressor} model (first row of Fig.~\ref{fig:compressor}).
Compared to the baseline optimization with ray-tracing, it is evident that the iso-points augmented optimization consistently achieves better results  at every timestamp. In other words, 
with iso-points we can reach a given quality threshold faster.

\vspace{-0.5mm}
\section{Conclusion and Future Work}
In this paper, we leverage the advantages of two interdependent representations: neural implicit surfaces and iso-points, for 3D learning. Implicit surfaces can represent 3D structures in arbitrary resolution and topology, but lack an explicit form to adapt the optimization process to input data. 
Iso-points, as a point cloud adaptively distributed on the underlying surface, are fairly straightforward and efficient to manipulate and analyze the underlying 3D geometry. 

We present effective algorithms to extract and utilize iso-points. Extensive experiments show the power of our hybrid representation. We demonstrate that iso-points can be readily employed by state-of-the-art neural 3D reconstruction methods to significantly improve optimization efficiency and reconstruction quality. 

A limitation of the proposed sampling strategy is that it is mainly determined by the geometry of the underlying surface and does not explicitly model the appearance.
In the future, we would like to extend our hybrid representation to model the joint space of geometry and appearance, which can in turn allow us to apply path-tracing for global illumination,
bridging the gap between existing neural rendering approaches and classic physically based rendering. 
Another promising direction is utilizing the new representation for consistent space-time reconstructions of articulated objects.

\section*{Acknowledgement}
We would like to thank Viviane Yang for her help with the point2surf code. This work was supported in parts by Apple scholarship, SWISSHEART Failure Network (SHFN), and UKRI Future Leaders Fellowship [grant number MR/T043229/1].
\clearpage

{\small
\bibliographystyle{ieee_fullname}
\bibliography{DSS2_ref}
}

\end{document}